# On applying Neuro- Computing in E-com Domain

Asif Perwej
asifperwej@gmail.com
Singhania University, Rajsthan, India

**Abstract**
Prior studies have generally suggested that Artificial Neural Networks (ANNs) are superior to conventional statistical models in predicting consumer buying behavior. There are, however, contradicting findings which raise question over usefulness of ANNs. This paper discusses development of three neural networks for modeling consumer e-commerce behavior and compares the findings to equivalent logistic regression models. The results showed that ANNs predict e-commerce adoption slightly more accurately than logistic models but this is hardly justifiable given the added complexity. Further, ANNs seem to be highly adaptive, particularly when a small sample is coupled with a large number of nodes in hidden layers which, in turn, limits the neural networks' generalisability.

## 1. Introduction

During recent years, researchers have utilized Artificial Neural Networks (ANNs) to solve a wide array of business and marketing research problems including regression [7, 12], classification [29, 31] and clustering/grouping, with some promising results. The ANNs seem to be particularly amenable to classification problems where the outcome variable encompasses two or more categories [30]. Earlier studies which compared the predictive performance of discriminate analysis with that of neural networks all concluded that ANNs are superior analytical tools when dealing with classification problem [28, 29, 31]. More recently, Shih and Fang (2005) [27] observed that ANNs were more accurate than both Discriminate Analysis and Decision Tree Analysis in predicting online shopping intention.

Despite all the favorable findings, the ANNs' literature remains mixed as there is also some discouraging evidence (see for e.g., [15]). Given the inconsistency in the literature, this paper intends to provide further empirical evidence by comparing logistic regression and neural networks in the context of consumer e-commerce. The remainder of the paper is organized into the following four sections. First we review the consumer e-commerce literature followed by a section on developing three ANNs. The third section discusses the performance of neural networks and compares it to logistic regression. Finally, the concluding section highlights some key findings emerged from our analysis.

## 2. Related Research Review

In seeking to explain the adoption of e-commerce by consumers, researchers have employed various theories and conceptual frameworks such as the Technology Acceptance Model [23], domain-specific innovativeness [11], shopping orientation [4], [9] and consumers' perceptions of online risk [10,19]. Demographic variables have also been widely used to predict consumer adoption behavior. Past results indicate that adopters of e-commerce tend to be young [8, 13]. Also, previous research examining the effect of sex revealed that males are more likely to conduct online transactions than females [4, 8, 16, 17, 20, 14]. It is also found that household income positively affects adoption of online shopping [17], had a positive impact on frequency and amount spent online [6, 16, 20] as well as predicting intention to shop via the internet [14]. In addition, empirical findings indicate that better educated individuals are more likely to favour online shopping [14, 17], purchase more frequently and spend more online [6].

The usefulness of demographics as predictor of consumer buying behavior has also been tested in context other than e-commerce with promising results [1, 5, 18 22, 24, 25]. Therefore, based on the past empirical findings, the current study uses nine demographic variables, cited by [26] as the most commonly used demographics, to predict adoption of consumer e-commerce.

## 3. Implementing neural networks

Essentially, artificial neural network is computer software which attempts to mimic biological neural systems. Like its biological counterpart, ANN has ability to learn complex patterns from data and subsequently generalize them into new contexts. It is suggested that ANNs are particularly useful when data exhibit non-linear and interaction relationships [15]. Details of ANNs are beyond the scope of this paper, however, for an excellent non-technical discussion, consult [21, 30].

Three "feed forward" Multi-layer Perceptron Networks [3] were developed to discriminate the adopters of consumer e-commerce from the non-adopters. The Networks share the same input and output layer but differ in terms of the number of processing elements (1-node, 5-node and 10-node network) in the hidden layer. Figure 1 illustrates the network with 10 nodes in the hidden layer. There are nine nodes in the input layer which correspond to the nine independent demographic variables. Previous studies suggested that the number of processing elements in the hidden layer is positively associated with the predictive performance of the networks [3]. However, there is a downside to this. A greater number of nodes in the middle layer increases the so-called "network capacity" which, in turn, limits the network's generalisability. It was mainly due to this reason, as well as gaining an in-depth understanding of ANNs' behavior under different circumstances, that we fitted three different types of networks to the data (i.e., with one, five and ten nodes in the hidden layer). The output layer consists of two nodes, one being for adopters and the other for non-adopters which essentially are the same as outcome variables in the logistic regression analysis. Finally, the epoch or number of iterations of the networks was set to 200 which is the maximum iteration available in our software (*i.e.,* Nuclass 7.06)

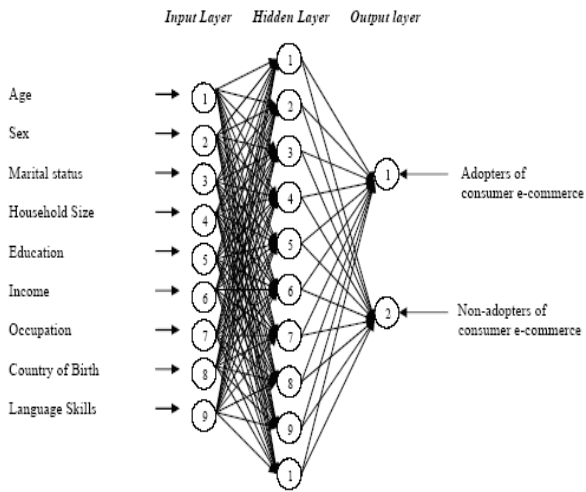

Figure-1 Neural architecture for adoption of consumer behavior

## 4. Results and discussions

Data for this study are drawn from General Social Survey 2002 (Australian Bureau of Statistics, 2003). The dataset contains information on 15,510 individuals' actual e-shopping, e-government, e-banking and e-share trading adoption which were treated as dependent variables and coded as 1= adopter and 2 = non-adopter. To avoid unequal cell size problem, we have created 20 sub-samples with an equal number of adopters and non-adopters (this later one selected randomly from corresponding non-adopter respondents) for each sample.

Table 1 depicts the results of Artificial Networks. Two major findings have emerged form this analysis. First, the networks exhibit good predictive power regardless of the number of processing elements in the hidden layer. The hit ratio ranges from 67.1 percent to 92 percent which should be compared to the Maximum Chance Criterion (MMC) of about 50 percent. Second, the networks with higher number of nodes in the middle layer seem to perform better then the networks with lesser nodes in the hidden layer. For example, on average, the networks with 10 nodes have correctly classified 78 percent of the respondents in terms of adoption of consumer e-commerce whereas comparable figure for the networks with five and one node in the hidden layer were 76 and 72 percent, respectively. **Comparing Results of Artificial Neural Networks and Logistic Models** ANOVA analyses were performed to compare the results of the Neural Networks with performance of logistic regression analysis which were reported by [24]. Note that, to facilitate comparability of the results, the same 20 samples which were used by [24] for logistic regression analysis were employed to conduct the ANN analysis.

Table-1

| Ref. | Dependent variables | Sample size | Hit Ratio | | | Maximum chance criterion | Logistic regression Hit Ratios |
|---|---|---|---|---|---|---|---|
| | | | With 1 node in hidden layer | With 5 node in hidden layer | With 10 node in hidden layer | | |
| 1 | e-banking | 7124 | 69.9 | 71.2 | 71.7 | 50.3 | 70.5 |
| 2 | e-share trading | 988 | 73 | 73.9 | 76.5 | 50.2 | 74.5 |
| | **Online Shopping** | | | | | | |
| 3 | Food and grocereries | 575 | 72.9 | 76.2 | 80 | 50.3 | 73.9 |
| 4 | Alcohol | 188 | 78.2 | 86.7 | 92 | 50.5 | 85.1 |
| 5 | Toys | 253 | 72.7 | 81.8 | 80.2 | 50.2 | 75.9 |
| 6 | Videos or Dvd's | 547 | 73.9 | 77.7 | 79.5 | 50.3 | 73.3 |
| 7 | Music or CD's | 786 | 68.3 | 75.7 | 76.7 | 50.5 | 71.8 |
| 8 | Books or Magazines | 1445 | 70.4 | 72.8 | 74 | 50.2 | 69.9 |
| 9 | Financial services | 678 | 73.2 | 76.1 | 79.4 | 50.7 | 73.3 |
| 10 | computer hardware or peripherals | 446 | 76.9 | 80.5 | 83.6 | 50.4 | 75.3 |
| 11 | Computer software | 997 | 69.9 | 72.6 | 73.8 | 50.1 | 69.9 |
| 12 | Clothing or shoes ,etc | 675 | 71.1 | 74.8 | 77 | 50.1 | 73.6 |
| 13 | Sporting equipment | 277 | 76.2 | 79.8 | 81.6 | 50.5 | 75.3 |
| 14 | Travel or Accommodation | 2412 | 72.4 | 73.3 | 73.8 | 50 | 73 |
| 15 | Ticket to entertainment or the cinema | 1217 | 73.2 | 77.2 | 76.2 | 50.5 | 74.5 |
| 16 | Other goods or services | 937 | 67.1 | 68.4 | 71.2 | 50.6 | 67.7 |
| | **E-government** | | | | | | |
| 17 | Electronic lodgment of tax returns | 1320 | 67.5 | 71.8 | 71.1 | 50.5 | 68.5 |
| 18 | Electronic lodgment of applications or claims for benefits | 310 | 72.6 | 78.7 | 81.6 | 50.3 | 74.8 |
| 19 | Electronic lodgement of application for permit:etc | 554 | 70.4 | 74.9 | 78 | 50.3 | 70.6 |
| 20 | Electronic lodgement of bill payments (e.g rate:and car registration ) | 3499 | 70.1 | 71.4 | 73.3 | 50.3 | 69.9 |
| | **Mean** | | 72 | 76 | 78 | 50 | 73 |

Table-2

| | Mean | Standard Deviation | f-value | p-value |
|---|---|---|---|---|
| Logistic regression analysis | 73.13 | 3.82 | | |
| Artificial Neural Network Analysis | | | | |
| Hidden layer with 1 node | 71.99 | 2.95 | 1.095 | 0.302 |
| Hidden layer with 5 nodes | 75.78 | 4.29 | 4.258 | 0.046 |
| Hidden Layer with 10 nodes | 77.56 | 5.01 | 9.901 | 0.003 |

Further, to be consistent with the results of the logistic models, all the data were submitted to training of the networks and, as such, no cross validation (*i.e.,* testing) was conducted. As can be seen from Table 2, no statistically significant discrepancy was detected between the result of logistic regression (mean hit ratio=73.13 percent) and the result of artificial neural network with one node in middle-layer (mean hit ratio=71.99 percent). However, the average hit ratio of the networks with five and ten nodes in the middle layer (75.78 percent and 77.56 percent, respectively) was significantly higher than mean hit ratio of logistic regressions.

Our findings corroborate the results of previous studies suggesting that neural networks may outperform conventional statistical models, such as logistic regression in this case. Two key findings emerged from this study. First, the increased accuracy is marginal and may not outweigh higher complexity involved with designing and operating ANNs. In the best case scenario, the ANNs added only about 4.5% to the accuracy of logistic models. Second, the performance of the neural networks seems to be affected by ANNs' architecture. We found, not surprisingly, that networks with iteration rate of 200 have yield better hit ratios than networks with, for example, only 50 iterations. Also the number of processing element in the middle layer exerted a positive influence on the networks' accuracy. The impact of sample size on the networks' performance is rather complex and not very well documented. Our observation indicates that when there is a limited number of patterns in place (*i.e.,* small sample size), the networks tend to memorise extant patterns amongst the data and consequently return higher predictive accuracy. Importantly, this particular characteristic of ANNs became more evident as the number of nodes in the hidden layer increases. As such, it would appear that small sample size magnifies the problem associated with including large number of processing elements in the hidden layer via increasing network capacity and adaptability.

No doubt, ANNs have brought about some great advances. Compared with conventional multivariate analysis, ANNs better handle multicollinearity, non-linear functions, noisy and missing data. One major problem with ANNs, among others, is the possibility of infinite network architecture in terms of network types, iterations, number of nodes and layers and so forth. This makes the outcomes of ANNs rather arbitrary. Given the indispensable role played by ANNs' architecture in determining the networks' outcomes, we suggest that future research endeavours in this field should focus on developing guidelines, similar to the "rules of thumb" used in conventional statistical methods. This will help researchers in the selection of the appropriate network features in order to obtain optimal and comparable model performance.